Paper Category: Original Article

Article Title: Comparison of CNN-based deep learning architectures for unsteady CFD acceleration on

small datasets

Authors: Sangam Khanal[1], Shilaj Baral[1], Joongoo Jeon[1,2*]

Full Credentials: Bachelor's degree, Bachelor's degree, Doctor of Philosophy

Job Title: Graduate student, Graduate student, Assistant Professor

Affiliations: [1]*Graduate school of Integrated Energy-AI, Jeonbuk National University*

[2]*Department of Quantum Systems Engineering, Jeonbuk National University*

*567 Baekje-daero, Deokjin-gu, Jeonju 54896, Republic of Korea*

*Corresponding author: Joongoo Jeon

Contact E-mail: jgjeon41@jbnu.ac.kr

Contact Phone: +82-2-2220-2355

Short Running Title: Comparison of CNNs for CFD

Word Counts: Abstract: 212, Main Text: 4,656

---

[*] Corresponding author: Email: jgjeon41@jbnu.ac.kr; Tel: +82-2-2220-2355

**Abstract**

CFD acceleration for virtual nuclear reactors or digital twin technology is a primary goal in the nuclear industry. This study compares advanced convolutional neural network (CNN) architectures for accelerating unsteady computational fluid dynamics (CFD) simulations using small datasets based on a challenging natural convection flow dataset. The advanced architectures such as autoencoders, UNet, and ConvLSTM-UNet, were evaluated under identical conditions to determine their predictive accuracy and robustness in autoregressive time-series predictions. ConvLSTM-UNet consistently outperformed other models, particularly in difference value calculation, achieving lower maximum errors and stable residuals. However, error accumulation remains a challenge, limiting reliable predictions to approximately 10 timesteps. This highlights the need for enhanced strategies to improve long-term prediction stability. The novelty of this work lies in its fair comparison of state-of-the-art CNN models within the RePIT framework, demonstrating their potential for accelerating CFD simulations while identifying limitations under small data conditions. Future research will focus on exploring alternative models, such as graph neural networks and implicit neural representations. These efforts aim to develop a robust hybrid approach for long-term unsteady CFD acceleration, contributing to practical applications in virtual nuclear reactor.

**Nomenclature**

| | |
|---|---|
| $u_x$ | Velocity in x-direction |
| $u_y$ | Velocity in y-direction |
| $t$ | Temperature |
| $\rho$ | Density |
| $u$ | Velocity vector |
| $p$ | Static pressure |
| $\alpha_{eff}$ | Effective Thermal diffusivity |
| $\mu_{eff}$ | Effective Viscosity |
| $h$ | Sum of internal energies |
| $K$ | Kinetic energy |

**Greek letters**

| | |
|---|---|
| $\alpha$ | Thermal diffusivity ($m^2$/s) |
| $\lambda$ | learning rate |
| $\rho$ | Density (kg/$m^3$) |

## 1. Introduction

Computational Fluid Dynamics (CFD) is critical in the analysis and simulation of fluid flows. It can provide tools to model complex phenomena like turbulence and buoyant flows with high accuracy. The ability to simulate these is integral to various industries including nuclear industry as it allows for the optimization of design processes, inevitably reducing the need for experimental setups [1]. CFD methods have the capability to model intricate three-dimensional flow and heat transfer characteristics in nuclear reactors and is likely to become increasingly significant in future reactor design and thermal hydraulics analysis [2,3]. Various techniques like finite difference [3], finite volume [4], finite element have been utilized for CFD simulations, but they require significant processing power making unable to do real-time simulation despite boasting higher accuracy.

Artificial Intelligence (AI), especially deep learning (DL), has emerged as a promising technique across various scientific and engineering disciplines in recent years [5]. Its ability to leverage data to identify patterns has made it an invaluable tool for complex simulations and predictions. In the context of computational fluid dynamics, DL can be used to create models that bypass the computational burdens of traditional methods [6]. During training, the models leverage the parallel processing capabilities of Graphics Processing Units (GPUs) to handle computations efficiently. Once trained, these models can inference future time series with significantly reduced computational time outperforming numerical methods in terms of efficiency. Employing AI with CFD, leveraging novel architectures using convolutional neural networks (CNNs) and recurrent neural networks (RNNs) can significantly reduce computational costs and time. These advanced DL architectures can learn spatial information as well as temporal information of CFD time series data.

As shown in Table 1, DL has been vigorously employed in CFD throughout literature to decrease the computational costs. Representatively, Srinivasan et al. [7] used LSTM for vortex shedding flow highlighting the superiority of recurrent networks over traditional MLPs. Similarly, Stevens et. al [8] used LSTM architecture to mimic partial differential equations but was limited to only one spatial dimension. More recently, Shu et. al used DDPM for reconstruction of 2-dimensional Kolmogorov flow using a physics informed condition [9]. Primarily, two approaches are followed, first being using vanilla

AI techniques and secondly considering scientific knowledge such as governing equations [8–10].

Although many studies have reported that CFD inference accuracy and acceleration performance can be increased with advanced DL architectures, they each used different datasets. For proper comparison, it is essential to compare various neural network architectures on the same dataset leading to a concrete conclusion. Additionally, CFD simulations have a constraint that the simulations need to be autoregressive. Given a single timestep, the next $n$ timesteps need to be predicted recursively. This leads to an accumulation of errors as the predictions are prolonged. Even with the use of LSTM, Srinivasan.et.al highlighted that error accumulation began after 10 timesteps, progressively increasing thereafter. This is an inevitable issue in the traditional single training approach, which seriously undermines practicality.

To address this error accumulation issue, residual-based physics-informed transfer learning (RePIT) strategy, a DL and CFD hybrid solver, has recently been developed [25]. The key idea of this strategy is to cross-compute DL and CFD, so that the residual iteration in the CFD section addresses the error accumulation in the DL acceleration section. The reason why this cross-computation is feasible is that residuals of governing equations are monitored for every timestep including DL section. If the residuals of the predictions exceed a certain threshold, the simulation switches to using CFD solver to reduce the residuals by an iteration scheme. After the residual has recovered normally, the latest CFD data is then used for transfer learning on the DL, allowing it to predict the next time series. This iterative process of switching between DL and CFD allows for practical long-term unsteady CFD acceleration. Zhang et al. also demonstrated that a hybrid approach of DL and CFD is a practical method for stable CFD acceleration through the development of the hybrid iterative numerical transferable solver (HINTS). Since RePIT does not require costly pre-training of DL and can inference CFD future time series with only initial short time series, it is a universal solver whose usability does not depend on training data. However, its DL model should be able to accurately predict nearby time series under small dataset conditions. The finite volume method network (FVMN) used in their previous study is capable of inferring CFD with short time series, but their applications were limited to uniform grid conditions.

The objective of this study is to build advanced CNN architectures and compare their performance under the same small dataset. For dataset, we used natural convection dataset which is simple but challenging for DL. DL trained to minimize the average error over the whole domain, tend to converge to a trivial solution in natural convection dataset where the flow is concentrated only in the region near the walls. We believe that this advanced model comparison with fairness and practicality are important milestone toward the nuclear industry's CFD acceleration goal. It should be noted that CNNs are much more flexible to such geometry changes than FVMN because they process data in the form of images through convolutional kernel operations. Because many advanced CNN architectures such as autoencoder, UNet, and ConvLSTM have been developed, it is reasonable to examine the usability of these advanced architectures in the RePIT framework.

Section 2 introduces the fundamental concepts behind the building blocks of architecture used like convolution layer, transposed convolution, ConvLSTM and the backpropagation algorithm. Section 3 delves into the CNN-based architectures constructed for this study, describing the motivation behind each architecture with their key characteristics. The description of CFD dataset is explained in Section 4, detailing the case study of natural convection flow. Finally, Section 5 presents the results of applying various architectures to predict the fluid flow. We describe the two different strategies of inference, absolute value and difference value inference, and the metrics used to compare the performance of each model. Lastly, the discussion highlights the relative strengths of each approach. The codes of all the models introduced in this paper are available at the following address: https://github.com/JBNU-NINE/Compare_CNNs_forCFD.

## 2. CNN Layers

Images exhibit unique spatial properties explained by two assumptions: locality of pixel dependencies and stationarity of statistics [26]. The first assumption highlights that nearby pixels share more related features, influencing each other more than distant pixels. Assuming similar statistics in local patches allows applying the same convolutional filters across various locations for feature extraction. If a feature appears in different locations, these filters effectively capture similar features. On the other hand, the second assumption indicates that the statistical properties of local image patches remain consistent across different locations and even different images.

These properties enables convolutional filters to operate on small, localized regions, understanding the relationships among neighboring pixels and contributing to the formation of local features. Ultimately, due to this focus on locality, fewer parameters can be used, resulting in higher efficiency compared to fully connected networks. Convolutional neural networks are known for taking advantage of spatial information outperforming other techniques in data with 2D, 3D shapes [27]. CNNs have also been widely used in fluid dynamics to exploit the spatial information between the fluids to simulate fluid flow [16,28]. In this section, we introduce different convolution layers that were used in creating our model architectures.

## 2.1 Convolution

In this study, we build various CNN models and evaluate their performance. First, this section introduces the principle of convolution operation, widely known in signal processing. The convolution represents the amount of overlap between two signals when one is reflected about y-axis and shifted, mathematically written as:

$$(f * g)(t) := \int_{-\infty}^{\infty} f(\tau)g(t-\tau)d\tau \qquad\qquad 1$$

Writing **Eq. (1)** in discrete two-dimensional form for a two-dimensional input $I$ and a two-dimensional kernel $K$ and using the commutative property of convolution as follows:

$$S(i,j) = (K * I)(i,j) = \sum_m \sum_n I(i-m, j-n)K(m,n) \qquad\qquad 2$$

In **Eq. (2),** the kernel flipping causes the commutative property. In the case of neural networks, this property is not usually important [29]. So, the equation can be changed, calling the function cross-correlation as follows:

$$S(i,j) = (K * I)(i,j) = \sum_m \sum_n I(i+m, j+n)K(m,n) \qquad 3$$

In machine learning, this equation which implements cross-correlation is commonly known as convolution which is shown in **Figure 1(a).** Finally, gradient descent can be used to learn the values inside the filter (weights) that minimize the loss function [30]. Furthermore, as convolution is a linear operation, non-linear activation functions are added in conjunction to approximate non-linear functions [31].

## 2.2 Transposed Convolution

This section introduces transposed convolutions, also known as fractionally strided convolutions which are primarily used for upsampling in autoencoders. Transposed convolutions swap the forward and backward passes of convolution operation. The operation is defined by the kernel and the way of computation of forward or backward pass determines it being convolution or transposed convolution [32]. If the input and output were unrolled into vectors in a left-to-right, top-to-bottom order, the convolution operation could be represented as a sparse matrix $C$ [32] as shown in **Figure 1(a)**. Then, the convolution can be represented where $I_c$ is the input and $O_c$ is the output as:

$$O_c = C.I_c \qquad 4$$

Taking the transpose of convolution matrix, the shapes of O and I will change, therefore Transposed Convolution where $I_{CT}$ is the input and $O_{CT}$ can be represented as:

$$O_{CT} = C^T.I_{CT} \qquad 5$$

As shown in **Figure 1(b)**, the transposed convolution can be represented with the transpose of a convolution matrix, where the input and output shapes are interchanged when compared with regular convolution.

## 2.3 ConvLSTM

The Convolution based architectures only use the spatial features in the dataset. In a fluid simulation, the features are also temporally related with each other i.e. variable fields on $t_{n+1}$ is not dependent only on variable fields on $t_n$ but also past variable fields beyond that [33]. To exploit this, a spatio-temporal filter, ConvLSTM can be employed. ConvLSTM is a layer that replaces the matrix multiplication in LSTM with convolution operation to capture spatial as well as temporal features. LSTMs are gated RNNs that have been known to handle long term dependencies [34]. Over the years, LSTMs have had many variants and LSTMs have been modified with convolution operation to create convolutional LSTM [35]. The ConvLSTM equations are shown below in **Eq. (6-11)** where '$\odot$' represents Hardmard product and '$*$' represents convolution operation. As shown in **Figure 2**, $X_t$ is the image input matrix.

$$i_t = \sigma(W_{xi} * X_t + W_{hi} * H_{t-1} + b_i) \qquad 6$$

$$f_t = \sigma(W_{xf} * X_t + W_{hf} * H_{t-1} + b_f) \qquad 7$$

$$o_t = \sigma(W_{xo} * X_t + W_{ho} * H_{t-1} + b_o) \qquad 8$$

$$g_t = \tanh(W_{xg} * X_t + W_{hg} * H_{t-1} + b_g) \qquad 9$$

$$C_t = f_t \odot C_{t-1} + i_t \odot g_t \qquad 10$$

$$h_t = o_t \odot \tanh(C_t) \qquad 11$$

By employing the above equations, the information in $H_{t-1}$ is also used in the calculation at the current state giving it the temporal property and is also called Short Term Memory. In turn, $C_t$ is used in conjunction with various gates to store and remove information from all timesteps and is also known as Long Term Memory as highlighted in **Figure 2**. It should be noted that, in this study, the matrix multiplications from the LSTM equations in **Eq. (6-9)** have been changed to convolution operations capturing the underlying spatial features as well as temporal features.

### 2.4. Backpropagation in CNNs

Backpropagation is the fundamental algorithm used to train neural networks, particularly in the context of DL [36]. It's an efficient way of computing the gradient of the loss function with respect to the weights of the neural network. This gradient is used to update the parameters in the direction that reduces the loss, a process commonly known as gradient descent [36]. The loss function $L(y', y)$ quantifies the difference between the predicted output by the neural network and the actual target output respectively [37]. The primary objective in neural network training is to minimize this loss function [37]. Different types of loss functions are used depending on the nature of the task [38,39] and in this case, since the output of the network is continuous, Mean Squared Error (MSE) is taken as the loss function.

The principles of backpropagation apply broadly across various network architectures, but the implementation in CNNs involves specific considerations due to their unique characteristics. Layers like convolutional and pooling layers introduce additional complexity in how gradients are calculated and propagated. The training process can be divided into two stages: forward pass and backward pass. The forward pass is where the network processes the input data to generate predictions as described in *Section 2.1*. In the backward pass, the gradients are calculated with respect to each layer and propagated backward through the network.

The gradient update rule in backpropagation where $\lambda$ is the learning rate, $w$ is the weights of the network and $L$ is the loss function is:

$$w_i' = w_i - \lambda \times \frac{\partial L}{\partial w_i} \qquad\qquad 7$$

Typically, the loss is not directly connected to the weights of the network. Therefore, the chain rule can be employed. For a CNN filter with $(3 \times 3)$ input and a $(2 \times 2)$ kernel and using **Eq. (3)**, the outputs can be written as:

$$z_1 = a_1 w_1 + a_2 w_2 + a_4 w_3 + a_5 w_4 \qquad 13$$

$$z_2 = a_2 w_1 + a_3 w_2 + a_5 w_3 + a_6 w_4 \qquad 14$$

$$z_3 = a_4 w_1 + a_5 w_2 + a_7 w_3 + a_8 w_4 \qquad 15$$

$$z_4 = a_5 w_1 + a_6 w_2 + a_8 w_3 + a_9 w_4 \qquad 16$$

Now, the partial derivative can be calculated where $y'$ is the predicted output using the chain rule:

$$\frac{\partial L}{\partial w_i} = \frac{\partial z_1}{\partial w_i} \times \frac{\partial L}{\partial z_1} + \frac{\partial z_2}{\partial w_i} \times \frac{\partial L}{\partial z_2} + \cdots \qquad 17$$

Using **Eq. (12) and (17)**, the partial derivative $\frac{\partial L}{\partial w_i}$ can be calculated for each value of $i$ and the weight update can be calculated where the operations are same to the transposed convolution in **Figure 1(b)** except the gradients are calculated in this case.

## 3. CNN Models

As noted in *Section 1*, we constructed various CNN models with different network architectures to compare the accuracy based on the 2D natural convection simulation. In this section, we describe the motivation behind each architecture outlining their key characteristics.

### 3.1 Autoencoder

An autoencoder consists of two main components: an encoder and a decoder as shown in **Figure 3(a)**.

The encoder maps the input data to a lower-dimensional latent space, capturing the most significant features and patterns within the data. The decoder then reconstructs the original data from this latent representation [40]. The strength of autoencoder lies in its ability to learn an efficient representation of the data that preserves essential information. The latent space representation is particularly effective because it allows the model to filter the noise and focus on the most informative aspects of the input [41]. In fluid mechanics, the ability to distill complex flow patterns into compact representations in the latent space is beneficial not only for reconstructing the original state but also for predicting subsequent states with high fidelity [42]. Therefore, our investigation into the performance of autoencoders is driven by their ability to generate compact yet detailed representations, which are crucial to achieve reliable predictions in dynamic systems.

The three properties from the dataset—velocity in the x-direction($u_x$), velocity in the y-direction($u_y$), and temperature($T$)—are taken as inputs to the neural network. Initially, a convolutional downsampling operation is performed by the encoder, followed by convolutional upsampling by the decoder, finally inferring the same variables $(u_x, u_y, T)$ at the next time step. The encoder applies $3 \times 3$ convolutional filters, followed by $2 \times 2$ pooling to reduce the features. These filters are applied three times, progressively reducing the spatial dimensions of the features. The specific change in dimensions depends on the size of the convolutional kernels, pooling windows, and the stride used during these operations. For instance, using a larger stride or pooling window will result in a more significant reduction in feature dimensions [43].

After consequent convolution and pooling operations, the features are encoded into a latent space with reduced dimensionality. The decoder then applies up-sampling and transposed convolution filters to restore the features from the latent space, reconstructing the original input dimensions. Finally, a sigmoid function is applied to constrain the output values between 0 and 1. After de-normalization, the next variable field prediction $(u_x, u_y, T)$ is completed.

### 3.2 UNet

The UNet architecture as shown in **Figure 3(b)**, originally proposed for brain segmentation [44], is an encoder-decoder model enhanced with skip connections that directly link each layer in the encoder with its corresponding layer in the decoder. The skip connections are crucial as they mitigate information loss for precise feature localization, which is particularly useful in tasks requiring fine-grained spatial information[45]. Additionally, the skip connections alleviate the vanishing gradient problem, enabling the network to train more effectively with deeper architectures[45].

Our application involves prediction and reconstruction, maintaining high resolution spatial details to accurately capture the complexity of fluid flow. Unlike traditional autoencoders, the symmetrical structure of UNet, with skip connections provides significant advantages in preserving and reconstructing these details. By experimenting with different UNet variants, such as adjusting padding and network depth, we aim to optimize its performance for our specific needs. These modifications leverage UNet's inherent ability to retain spatial information, making it a valuable architecture for fluid mechanics applications.

### 3.2 ConvLSTM-UNet

The original UNet architecture faced a significant limitation in its inability to extract temporal information, relying solely on spatial data. However, for fluid simulations, capturing temporal dependencies is crucial, as the state of fluid at time $(t + 1)$ depends on its state at times $(t, t - 1, t - 2, ...)$. RNNs and LSTM can effectively capture temporal information [46], but conversely they are limited to utilize spatial information.

In the UNet architecture discussed in the previous section, we propose incorporating a ConvLSTM layer at the input stage as in **Figure 3(c)**. This layer accepts input in the format of (timesteps, channels, height, width), where each timestep represents the fluid state. The memory cell $C_t$ extracted from the ConvLSTM layer after processing all timesteps encapsulates spatio-temporal information from the entire sequence. This integration of spatial and temporal features through ConvLSTM will enhance the model's ability to capture complex fluid dynamics phenomena. The codes of all the models introduced in this paper are available at the following address: https://github.com/JBNU-NINE/Compare_CNNs_forCFD.

**4 CFD Dataset**

In exploring the viability of neural networks in fluid dynamics, we used a 2D natural convection flow CFD dataset used in a previous RePIT study[25]. Although it is a challenging dataset with three governing equations (mass, momentum and heat transfer), dealing with heat transfer equations is very important and meaningful in CFD applications. The vertical sides of the grid were a hot wall (307.75K) and a cold wall (288.15K). The two horizontal walls were adiabatic walls. Using this, a mesh was created of 40,000 uniform cells. The initial temperature inside the grid was set to 293K and the pressure was 1 atm. Using the above conditions, a simulation of 100s was conducted with a step of 0.01 seconds (10,000 timesteps).

The BuoyantPimpleFoam solver from the OpenFOAM library was used to solve the fluid equations[47]. BuoyantPimpleFoam is a transient solver that combines the PISO and SIMPLE algorithms, making it suitable for buoyancy-driven flows involving heat transfer. It employs the Boussinesq approximation [48] to account for density variations due to temperature differences, which drives the natural convection in the system. The fluid is driven by buoyancy and gravity. When air near the hot wall gets heated, it rises due to buoyancy and is transported towards the cooler wall through the adiabatic upper wall. As the air molecules hit the cold wall, they exchange heat, causing the density to increase and the air to be drawn downward by gravity. The solver effectively handles the incompressible Navier-Stokes equations, continuity equation, and energy equation, ensuring accurate simulation of the thermal and fluid flow behavior in the grid [48]. As the flow begins to stabilize over time, the time range 0-10 s were taken for the comparative study of this paper (**Figure 4**). Because DL trained to minimize the average error over the whole domain, tend to converge to a trivial solution in natural convection dataset where the flow is concentrated only in the region near the walls. Additionally, the DL model never saw the fluid states of test dataset during training procedure (generalization). In other words, the natural convection dataset is challenging for DL. Compared to the literatures on CNN applications, an image dataset if this scale (1,000) falls under the small dataset condition, which fits the usability investigation in RePIT.

## 5 Results and Discussion

We applied the various architectures discussed in the previous sections in two approaches: absolute value calculation and difference value calculation. We adopted these methods because difference calculations are naturally used in traditional CFD solvers to capture dynamic changes, while convolutional networks like autoencoders are effective at modeling spatial patterns and reconstructing entire flow fields. In the first approach, the field data at timestep $t_n$ is input to predict the field data at next timestep $t_{n+1}$. In the latter approach, the difference between flow field data at timestep $t_n$ and $t_{n+1}$ was predicted by the input of field data at timestep $t_n$. Using these two strategies, the neural networks were trained for the first 800 timesteps. The next hundred (801-900) timesteps were taken as validation data. Finally, all the predictions were done for the next 100 timesteps, i.e. 901-1000 timesteps. This ensured that the model never saw the fluid states in which the predictions would be made.

The 901st timestep fluid state was given as the input in the case of autoencoder and UNet. In the case of ConvLSTM-UNet, 901st to 905th timestep fluid states were given as inputs. All other states were recursively predicted. The inputs to the model were normalized between 0 and 1 and consequently the outputs from the model are also in the same range. The metrics were calculated after de-normalizing the outputs of the network to original value ranges. To precisely analyze the model outputs, the maximum error value over the entire grid was taken as the primary metric for calculation. Additionally, for comparison with the regressive generation, sequential generation where the model is given a fluid state and the model predicts the only one subsequent state was also investigated. In regressive generation, a starting fluid state is given to the model. After that, the model recursively predicts the future fluid states by feeding its outputs back as inputs like CFD solver.

### 5.1 Absolute Value Calculation

In the absolute value calculation, for the state of fluid at a given timestep $t_n$, its state at the next timestep $t_{n+1}$ is predicted. In this strategy, the network must recreate the fluid flow for the next timestep entirely. The maximum errors for the proposed architectures are shown in **Table 2** and visualized in **Figure 5**, where it is evident that the ConvLSTM-UNet model with regressive generation demonstrated the lowest maximum errors ($T = 3.61\ K, u_x = 0.11\ m/s, u_y = 0.13\ m/s$) compared to other models, indicating its higher accuracy. Conversely, the UNet model with regressive generation exhibited the highest maximum error $T$ value of 15.16, suggesting significant deviations. These results highlight the model's efficiency in predicting the fluid state entirely for the next timestep.

**5.2 Difference Value Calculation**

In this strategy, for the state of fluid at a given timestep $t_n$, the difference with the fluid state at next timestep $t_{n+1}$ is calculated and added to the values of $t_n$ to predict future time series. Here, the network leverages the fact that consecutive timesteps are just modifications of previous timesteps. The results are shown in **Table 3** and further visualized in **Figure 6**. ConvLSTM-UNet model with regressive generation showed much lower maximum errors ($T = 4.01\ K, u_x = 0.048\ m/s, u_y = 0.082\ m/s$), underscoring its accuracy and robustness. On the other hand, autoencoder with regressive generation showed highest maximum errors across all three parameters.

The ConvLSTM-UNet model consistently demonstrated superior accuracy and lower error margins across both calculation methods under the regressive generation type. In contrast, the autoencoder model exhibited the highest error values, highlighting its limitations in predictive accuracy. These findings provide valuable insights into the effectiveness of different models and error calculation methods. To verify the usability of the ConvLSTM-UNet model in the RePIT algorithm, the number of predictable future time series was investigated based on the maximum allowable error confirmed in the previous RePIT study. In previous studies, the CFD cross point was calculated based on the residuals, and the maximum allowable errors for each variable corresponding to the residuals were 0.4 K, 0.024 m/s, 0.024 m/s ($T, u_x, u_y$). As shown in **Figure 7**, it is possible to predict up to about 10, 40, and 30 future time step based on each variable value. Considering that the training dataset has 800 timesteps, this is not a satisfactory result. It seems that CNN-based DL models may not provide a practical solution for CFD acceleration. However, the efficiency can be increased in transfer learning phase in RePIT. This is our future work.

### 5.3 Residual Calculation

In CFD, the residuals indicate the local disparity of a conserved variable within each control volume. Consequently, each grid in the model possesses its individual residual value for every governing equation undergoing solution. The continuity equation, Navier Stokes equation for incompressible fluids and heat equation respectively are solved to calculate the residuals in the controlled volume. In an iterative solution, the residuals cannot be 0 but must fall below a certain threshold to confirm the solution convergence. The residual calculation equations are shown in **Eq. (18-20)**, where $\rho$ is density, $u$ is velocity vector, $p$ is the static pressure, $\alpha_{eff}$ is the effective thermal diffusivity, $\mu_{eff}$ is the effective viscosity, $h$ is the sum of internal energies and $K$ is the kinetic energy.

$$\frac{\partial \rho}{\partial t} + \nabla.(\rho u) = 0 \qquad\qquad 18$$

$$\frac{d\rho u}{dt} + \nabla.(\rho uu) = \nabla p - \rho g - \nabla.\left(\mu_{eff}(\nabla u + \nabla u^T)\right) + \nabla\left(\frac{2}{3}\mu_{eff}(\nabla.u)\right) \qquad\qquad 19$$

$$\frac{d\rho h}{dt} + \nabla.(\rho uh) + \frac{d\rho K}{dt} + \nabla.(\rho uK) - \frac{\partial p}{\partial t} - \nabla.\left(\alpha_{eff}\nabla h\right) - \rho u.g = 0 \qquad\qquad 10$$

For the best-performing model (ConvLSTM-UNet), we monitored the residuals to investigate the cause of the error accumulation problem. Three residuals—residual mass, residual momentum, and residual heat—were calculated for every timestep and compared with the actual values from the solver. Ideally, fluctuations in the residual values are not desired. Therefore, from Figure 8 residual values, it can be concluded that the difference calculation method has stable residuals when compared with absolute value calculation.

### 6. Concluding remarks

In this study, we compared various neural network architectures using the 2D natural convection dataset. The CNN-based methods that take only spatial features into account performed poorly compared to CNN+RNN-based methods (ConvLSTM-UNet). Furthermore, most of the details lie near the boundary region in the dataset. The use of CNN and Pooling could be a reason for the loss of information, especially from the boundary pixels. The results indicate that the difference value calculation method performs better than absolute value calculation from both perspectives: maximum error and residuals. This method leverages incremental changes between consecutive timesteps, resulting in lower errors. However, only 10 timesteps can be predicted within the accepted threshold. When the predictions are continued for longer timesteps, the errors become increasingly noticeable as highlighted in **Figure 9**. In the figure, the zones with most errors are boxed with red, which clearly diverge from the ground truth. Overall, these findings highlight the strengths and weaknesses of current CNN. While CNN and RNN-based methods show promise, significant improvements are needed, particularly in temperature prediction, to achieve longer and more accurate simulations. In other words, current image-based CNN architectures are difficult to achieve meaningful acceleration performance in transient CFD simulations.

To further enhance our model's capabilities, we plan to increase the training data size beyond the current 900 timesteps. Currently, the model accurately predicts only 10 of the next 100 timesteps, so increasing the training data should ideally enable it to predict more timesteps accurately. We will also explore the application of transfer learning after reaching an initial prediction threshold, utilizing actual data from CFD solvers like OpenFOAM to further refine the model. Additionally, we propose incorporating residuals into the loss function creating a physics informed loss function to minimize prediction errors, experimenting further to enhance overall accuracy and convergence behavior. Temperature prediction has been particularly challenging, with higher errors compared to other properties. To tackle this, we will modify the network in a decoupled way, separating temperature from other properties to focus on improving its accuracy. Exploration of other techniques like neural operator, graph neural networks (GNN) and implicit neural representation (INR) can also be done to address this challenge. Furthermore, we will integrate candidate models with the RePIT strategy, which checks residuals and switches to traditional CFD solvers as needed, ensuring accurate long-term predictions. Finally, we plan to release an open-source framework, including all the benchmarks, allowing for experimentation with various network architectures and integrating with the RePIT strategy. This will foster collaboration in the research community.

**Acknowledgments**

This research was supported by the National Research Council of Science & Technology (NST) grant by the Korea government (MIST) (No. GTL24031-300) and the Nuclear Safety Research Program through the Korea Foundation of Nuclear Safety (KoFONS) using the financial resource granted by the Nuclear Safety and Security Commission (NSSC) of the Republic of Korea (no. RS-2024-00403364).

**Figure captions**

**Fig. 1.** Convolution operations: a) normal convolution, b) transposed convolution

**Fig. 2.** ConvLSTM architecture

**Fig. 3.** Advanced CNN-based architectures: a) Autoencoder, b) UNet, c) ConvLSTM-UNet

**Fig. 4**. Evolution of temperature field in the natural convection simulation dataset (0 to 10 s). This dataset is suitable for evaluating the performance of advanced CNN architectures because the flow occurs only near the wall and there are grids where the temperature increase first starts at each timestep.

**Fig. 5.** Bar graph of the maximum error of absolute value calculation for temperature and x-velocity in test dataset

**Fig. 6.** Bar graph of the maximum error of difference value calculation for temperature and x-velocity in test dataset

**Fig. 7.** Maximum errors for absolute value and difference value calculation for temperature, x-velocity and y-velocity for autoregressive prediction of 100 timesteps (test dataset). Each point represents the thresholds (0.4, 0.024, 0.024) in a previous RePIT study.

**Fig. 8.** Residual mass, residual momentum and residual heat calculation for autoregressive prediction of 100 timesteps (test dataset).

**Fig. 9.** Comparison of predictions with CFD truth value for different timesteps for ConvLSTM-UNet architecture.

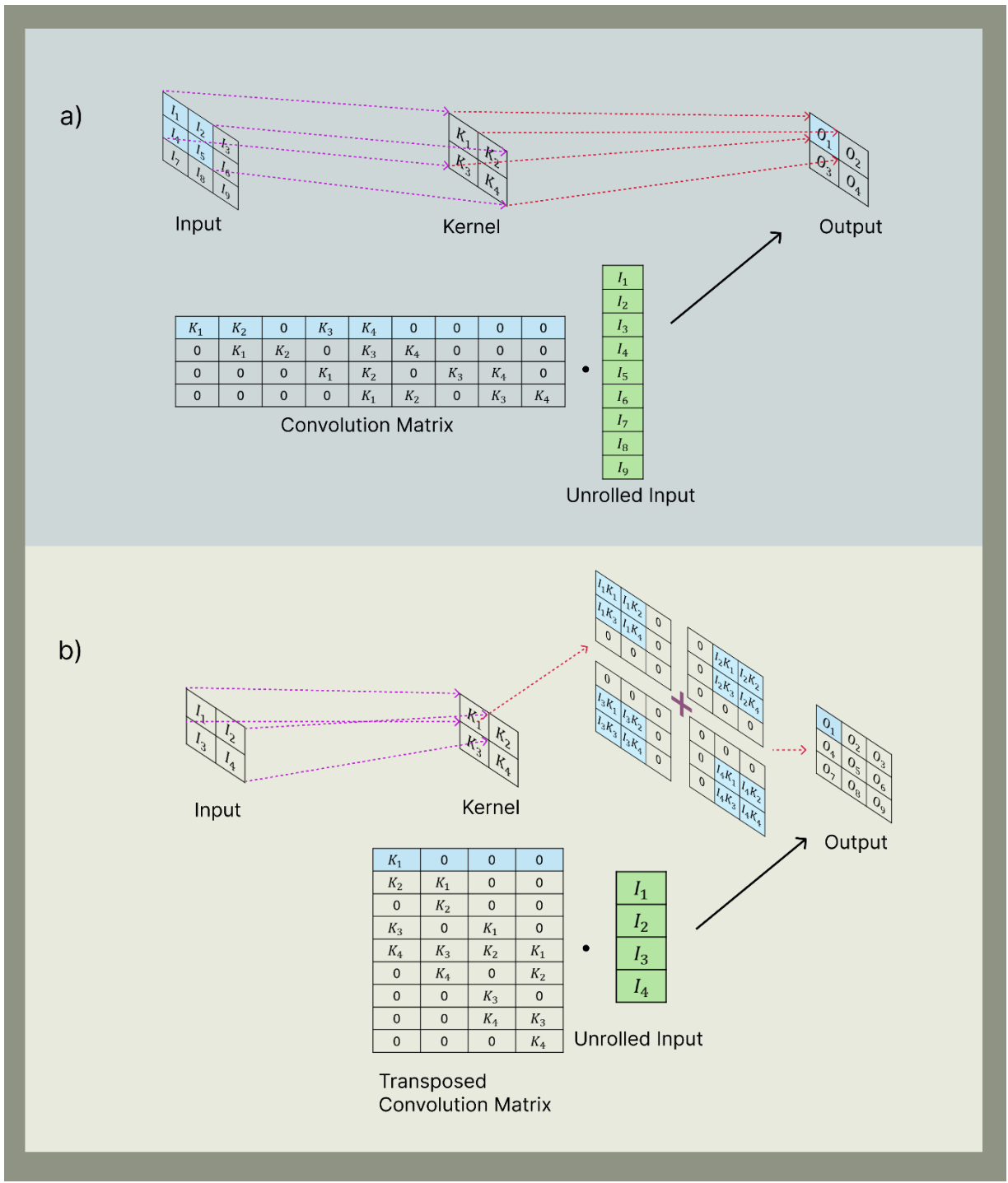

**Fig. 1.** Convolution operations: a) normal convolution, b) transposed convolution

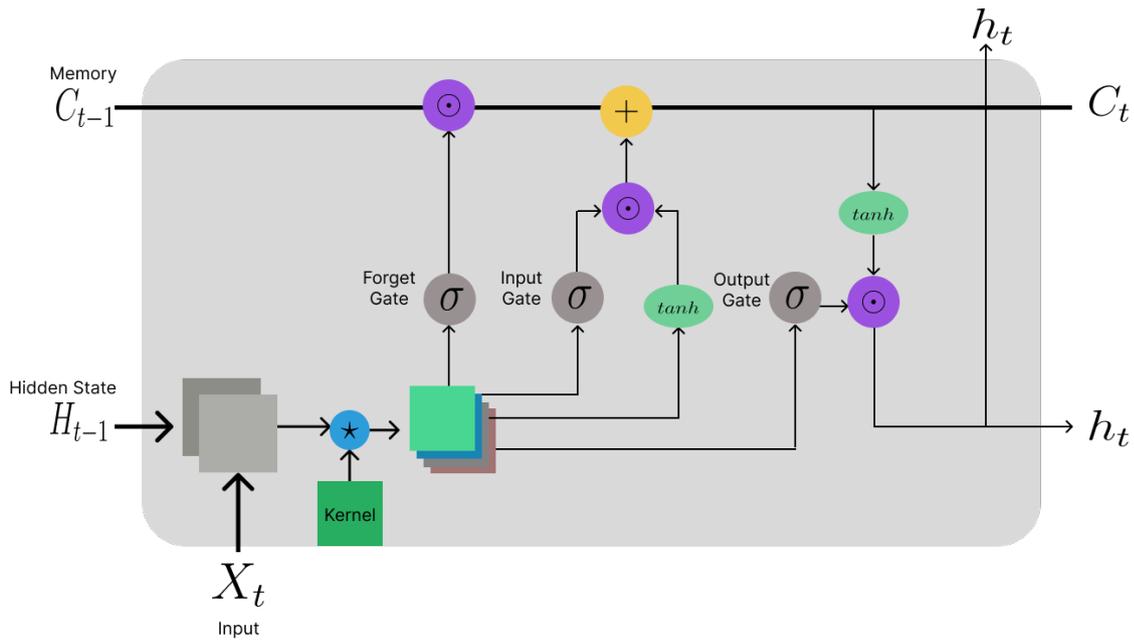

**Fig. 2.** ConvLSTM architecture

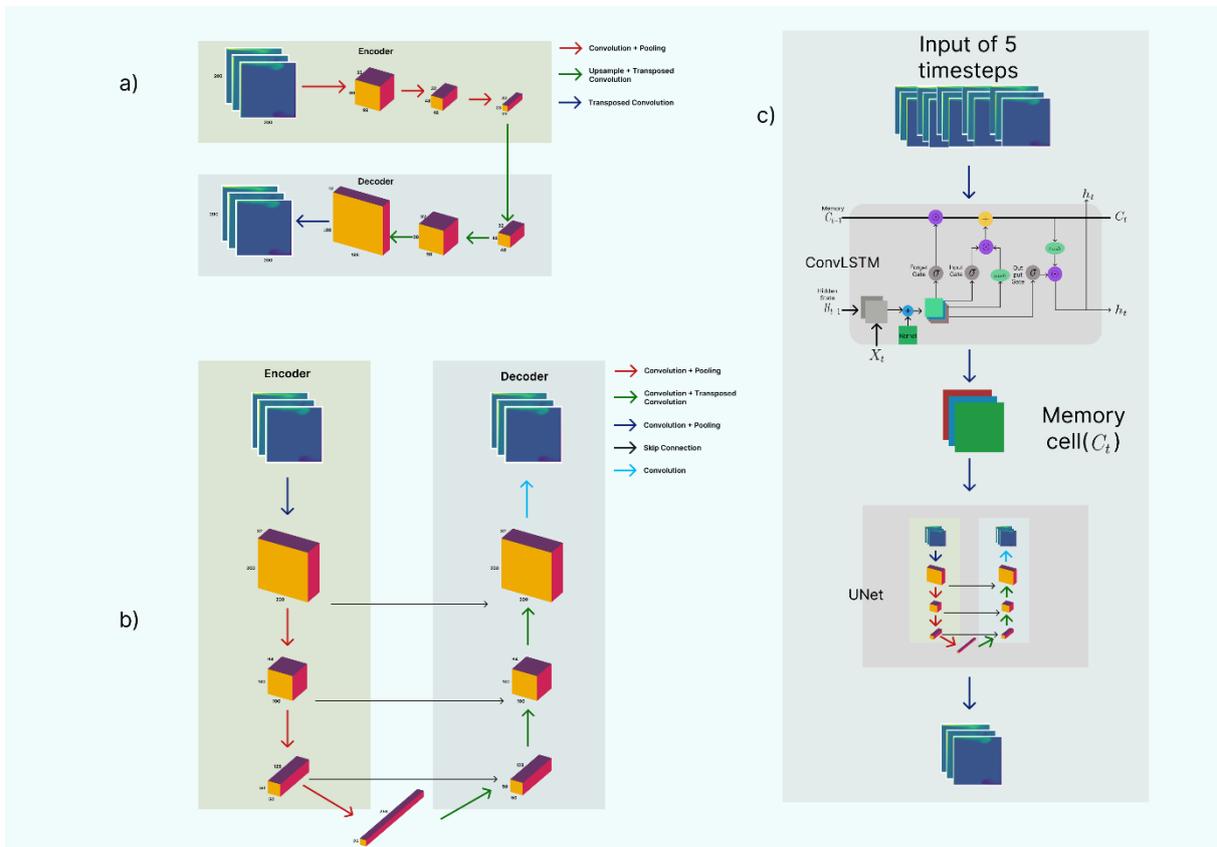

**Fig. 3.** Advanced CNN-based architectures: a) Autoencoder, b) UNet, c) ConvLSTM-UNet

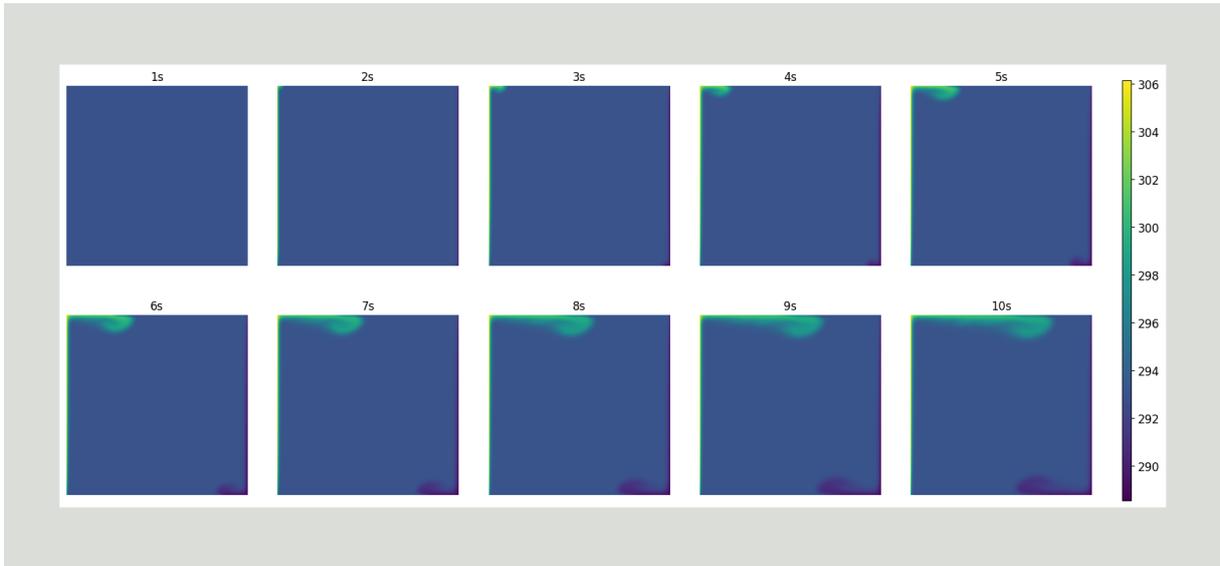

**Fig. 4.** Evolution of temperature field in the natural convection simulation dataset (0 to 10 s). This dataset is suitable for evaluating the performance of advanced CNN architectures because the flow occurs only near the wall and there are grids where the temperature increase first starts at each timestep.

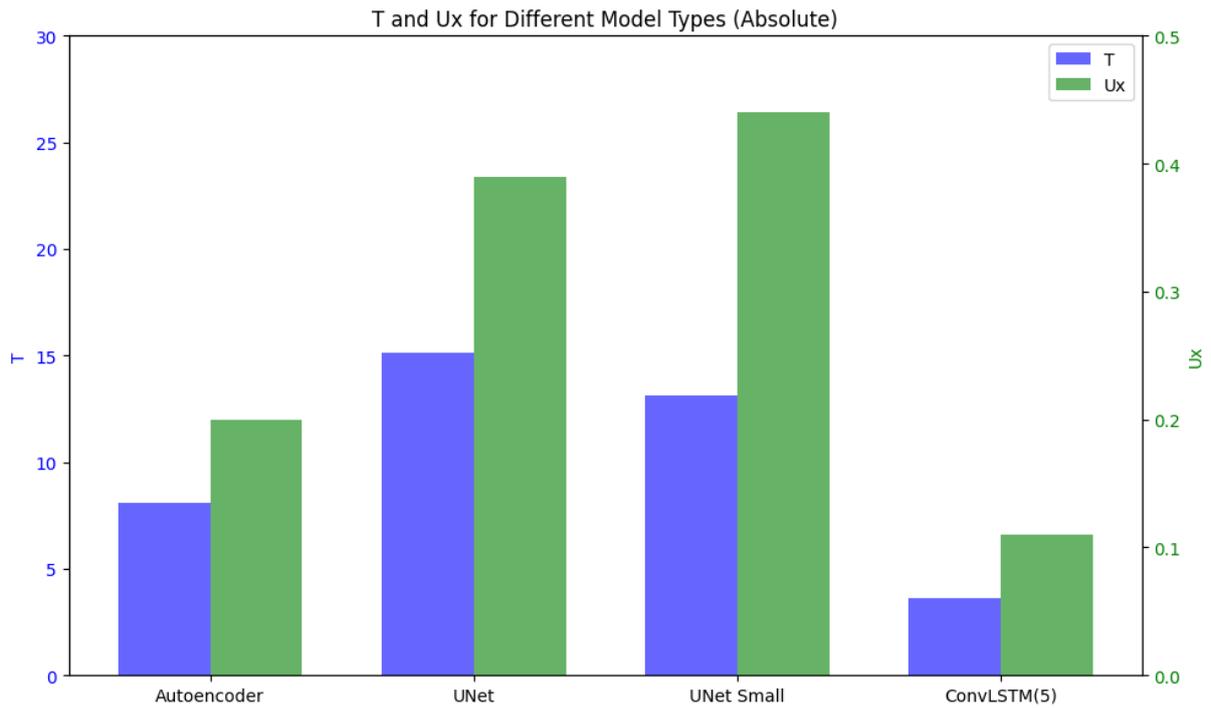

**Fig. 5.** Bar graph of the maximum error of absolute value calculation for temperature and x-velocity in test dataset

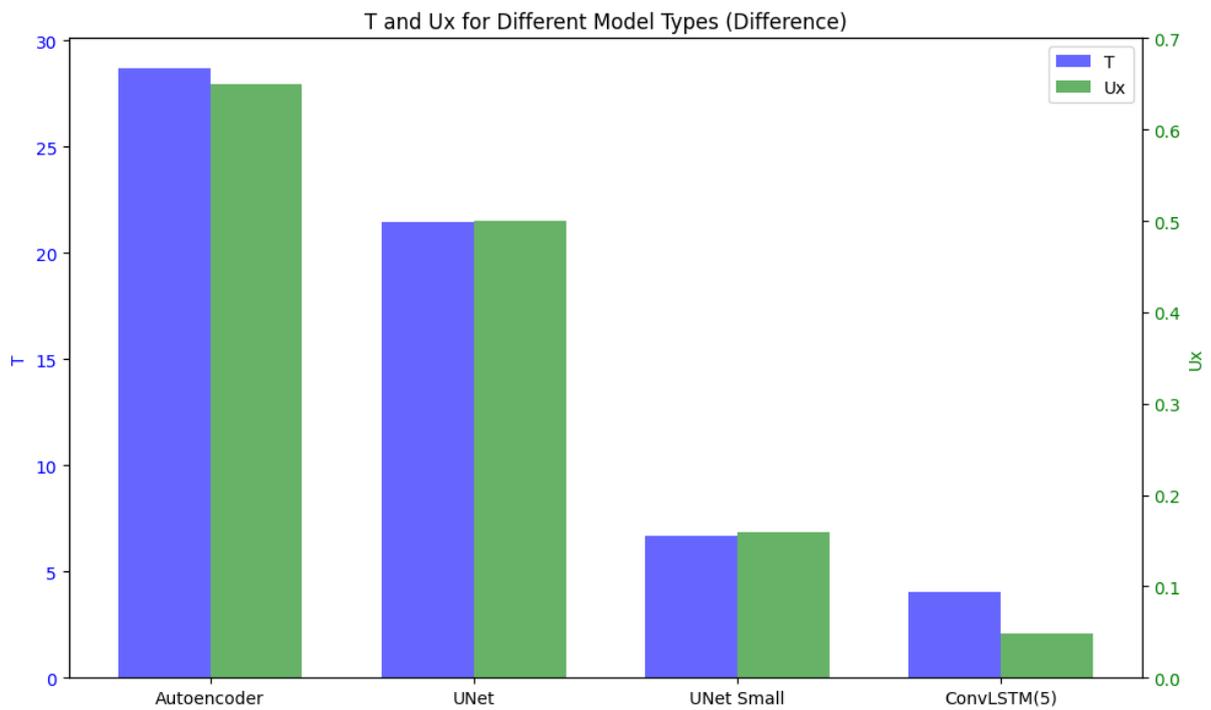

**Fig. 6.** Bar graph of the maximum error of difference value calculation for temperature and x-velocity in test dataset

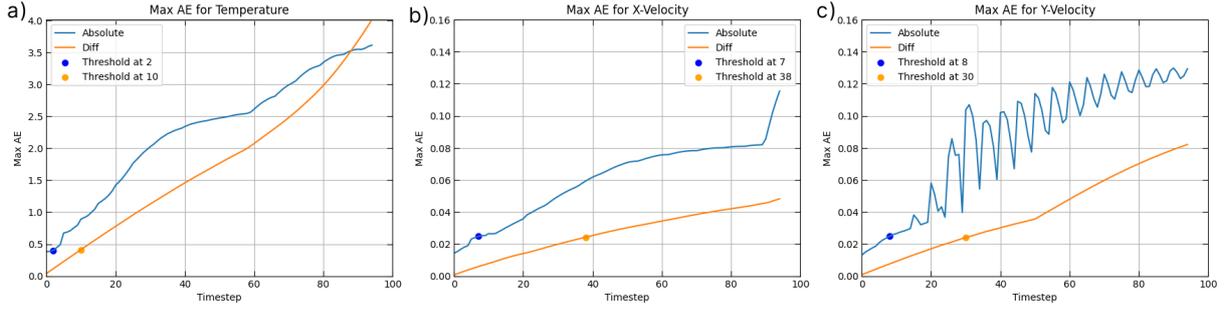

**Fig. 7.** Maximum errors for absolute value and difference value calculation for temperature, x-velocity and y-velocity for autoregressive prediction of 100 timesteps (test dataset). Each point represents the thresholds (0.4, 0.024, 0.024) in a previous RePIT study.

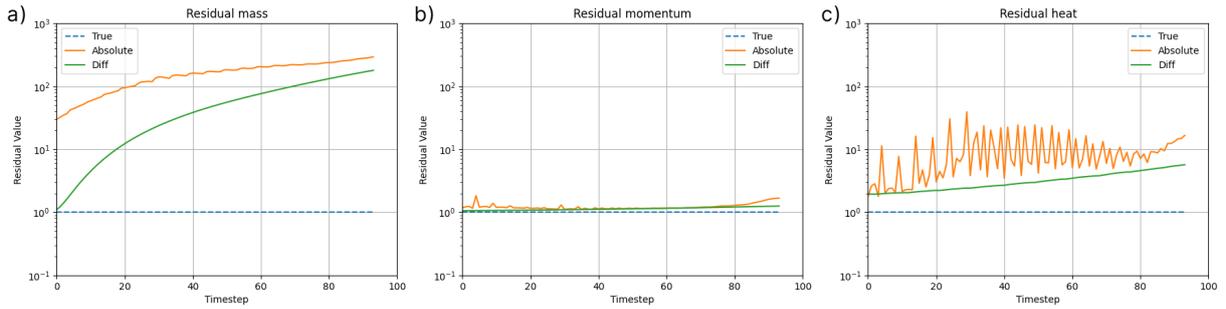

**Fig. 8.** Residual mass, residual momentum and residual heat calculation for autoregressive prediction of 100 timesteps (test dataset).

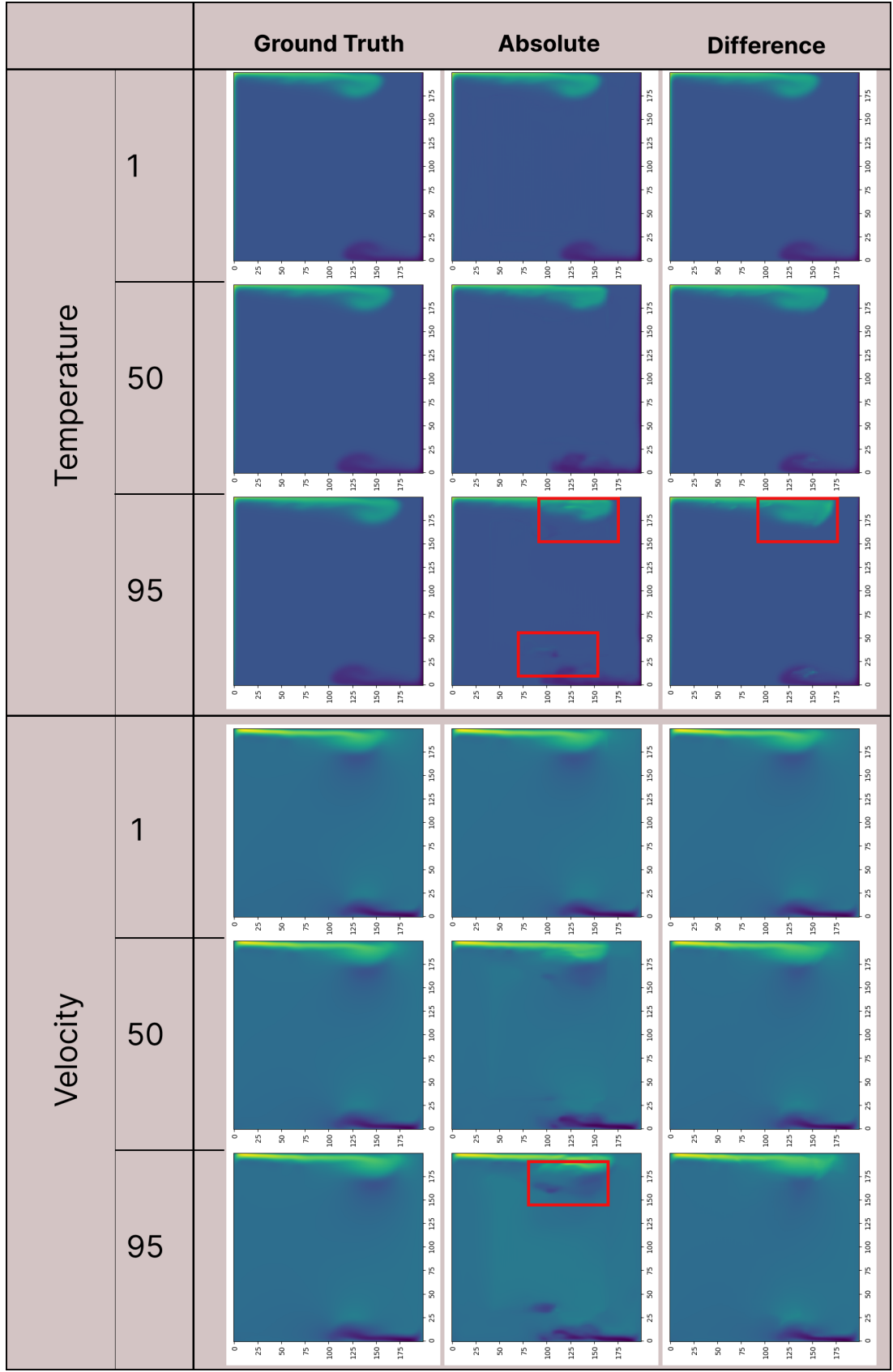

**Fig. 9.** Comparison of predictions with CFD truth value for different timesteps for ConvLSTM-UNet architecture.

**Table captions**

**Table 1.** Recent DL studies for CFD acceleration

**Table 2.** Absolute value calculation - maximum error for different architectures

**Table 3.** Difference value calculation maximum error for different architectures

**Table 1.** Recent DL studies for CFD acceleration

| Year | Author | Dataset type | Base network | Key ideas |
|------|--------|--------------|--------------|-----------|
| 2016 | Guo et al.[10] | Laminar flow | CNN | Steady laminar simulation |
| 2017 | Tompson et al.[11] | Euler equation | CNN | Solving Poisson equation |
| 2018 | San et al.[12] | Wind-driven ocean circulation | MLP | Reduced order modeling |
| 2019 | Lee et al.[13] | Vortex shedding flow | GAN | Physical loss function |
| 2019 | Srinivasan et al.[7] | Low-order model of turbulent shear flow | LSTM | MLP vs LSTM |
| 2019 | Beck et al.[14] | LES closure model | RNN | CNN-based RNN architecture |
| 2019 | Takbiri et al.[15] | Gas combustion | MLP | MLP with tier system |
| 2020 | Obiols-Sales et al.[16] | Wall bounded flow | CNN | DL coupled CFD for steady simulation |
| 2020 | Stevens et al.[8] | Inviscid Burgers's equation | LSTM | Convolutional layer corresponds to the stencil |
| 2020 | Ajuria et al.[17] | Incompressible flow | CNN | Solving Poisson equation |
| 2020 | Sun et al.[18] | Vascular flow | MLP | Physical constrained DL |
| 2021 | Kochkov et al.[19] | Incompressible N-S equation | CNN | Solving convective flux |
| 2021 | Eivazi et al.[20] | RANS equation | MLP | Solving spatial field with Physical loss function |
| 2021 | Praditia et al.[21] | Diffusion-sorption equation | MLP | Framework considering the FVM |
| 2021 | Ricardo et al.[22] | Various data type | DNN | Review article |
| 2022 | Jeon et al.[23] | Reacting/non-reacting flow | MLP | Finite volume method-based network design |
| 2023 | Kang et al.[24] | Laminar flow with obstacle | ViTransformer, | Couples ViT and U-Net |
| 2023 | Shu et al.[9] | 2D Kolmogorov flow | Diffusion | Flow field reconstruction |
| 2024 | Jeon et al.[25] | Natural convection | MLP | AI-CFD hybrid computation |

**Table 2.** Absolute value calculation - maximum error for different architectures

| Model Type | Generation Type | T | Ux | Uy |
|---|---|---|---|---|
| Autoencoder | Sequential | 5.53 | 0.05 | 0.09 |
| | Regressive | 8.09 | 0.2 | 0.1 |
| UNet | Sequential | 2.36 | 0.032 | 0.027 |
| | Regressive | 15.16 | 0.39 | 0.55 |
| UNet Small | Sequential | 1.85 | 0.027 | 0.53 |
| | Regressive | 13.1 | 0.44 | 0.5 |
| ConvLSTM-UNet | Sequential | 0.44 | 0.01 | 0.01 |
| | Regressive | 3.61 | 0.11 | 0.129 |

**Table 3.** Difference value calculation maximum error for different architectures

| Model Type | Generation Type | T | Ux | Uy |
|---|---|---|---|---|
| Autoencoder | Sequential | 0.28 | 0.006 | 0.024 |
| | Regressive | 28.67 | 0.65 | 2.46 |
| UNet | Sequential | 0.08 | 0.002 | 0.002 |
| | Regressive | 21.44 | 0.5 | 0.41 |
| UNet Small | Sequential | 0.079 | 0.001 | 0.002 |
| | Regressive | 6.67 | 0.16 | 0.26 |
| ConvLSTM-UNet | Sequential | 0.03 | 0.0007 | 0.0008 |
| | Regressive | 4.01 | 0.048 | 0.082 |